\documentclass[review]{elsarticle}

\usepackage{hyperref}

\usepackage{amssymb}
\usepackage{graphicx}
\usepackage{amsmath}
\usepackage{epsfig}
\usepackage{color}
\usepackage{booktabs}
\usepackage{multirow}
\usepackage{colortbl}
\usepackage{graphicx}
\definecolor{mygray}{gray}{.9}
\usepackage{caption} 
\usepackage{float}
\usepackage{gensymb}
\usepackage{subfigure}
\usepackage{diagbox}

\journal{Neural Networks}

\begin{document}

\begin{frontmatter}

\title{Deep Rival Penalized Competitive Learning for Low-resolution Face Recognition}





\author{Peiying Li}
\ead{lipeiying@sjtu.edu.cn}
\author{Shikui Tu\corref{mycorrespondingauthor}}
\ead{tushikui@sjtu.edu.cn}
\author{Lei Xu\corref{mycorrespondingauthor}}
\ead{leixu@sjtu.edu.cn}

\address{Department of Computer Science and Engineering, \\ Shanghai Jiao Tong University, Shanghai, China}
\cortext[mycorrespondingauthor]{Corresponding author}

\begin{abstract}
Current face recognition tasks are usually carried out on high-quality face images, but in reality, most face images are captured under unconstrained or poor conditions, e.g., by video surveillance. Existing methods are featured by learning data uncertainty to avoid overfitting the noise, or by adding margins to the angle or cosine space of the normalized softmax loss to penalize the target logit, which enforces intra-class compactness and inter-class discrepancy. In this paper, we propose a deep Rival Penalized Competitive Learning (RPCL) for deep face recognition in low-resolution (LR) images. Inspired by the idea of the RPCL, our method further enforces regulation on the rival logit, which is defined as the largest non-target logit for an input image. Different from existing methods that only consider penalization on the target logit, our method not only strengthens the learning towards the target label, but also enforces a reverse direction, i.e., becoming de-learning, away from the rival label. Comprehensive experiments demonstrate that our method improves the existing state-of-the-art methods to be very robust for LR face recognition. 
\end{abstract}

\begin{keyword}
Face recognition \sep Low resolution \sep RPCL
\end{keyword}

\end{frontmatter}


\section{Introduction} \label{sec:introduction}


Face recognition has been an active research topic in computer vision over decades \cite{Turk1991-eigen,Deepface_taigman2014deepface,DeepID_sun2014deep,DeepID2_sun2014deep}. It includes two sub-tasks: face identification and face verification under the open-set setting. Face identification classifies a face to a known identity, while face verification decides whether two input faces are from the same identity by measuring their similarity in the feature space. In real applications, the label space of the training set is usually different from that of the test set, and thus many efforts have been made to learn discriminant features for face recognition. 

Due to the recent development of deep convolutional neural networks (CNN), face recognition has achieved an unprecedented level of accuracy \cite{facenet_Schroff2015FaceNet,sphereface_liu2017,cosface_wang2018,arcface_deng2019,PFE_shi2019probabilistic,DUL_chang2020data}. One stream of methods is featured by the design of margin-based softmax loss, e.g., imposing multiplicative angular margin \cite{sphereface_liu2017} or additive margin \cite{arcface_deng2019} into the angle space, or additive margin into the cosine space \cite{cosface_wang2018}. The margin-based methods are able to effectively tackle the drawbacks of the original softmax loss, under which the learned features are separable for identities in the training set but not discriminative enough for testing identities out of the training set. There also exists a stream of research on learning embedding directly by maximizing face class separability, e.g., penalizing the softmax loss with the center loss \cite{Wen2016centerloss} for smaller intra-class variance, enlarging the Euclidean margin between triplet samples by the triplet loss \cite{facenet_Schroff2015FaceNet}. Another stream of research is to consider data uncertainty in face recognition \cite{PFE_shi2019probabilistic,DUL_chang2020data} by estimating each face image as a Gaussian instead of a fixed point, to avoid overfitting the noise.

The above methods achieve high accuracy in high-quality face images. However, when applying these methods to low-quality images, we usually see a significant decline in accuracy. In real-world applications such as video surveillance, faces in these images are generally in small size and low-resolution (LR), due to the poor environmental conditions, e.g., bad illumination, long-distance surveillance, and so on \cite{lipei2018low-survey}. To directly tackle the LR face recognition problem, LR images are often transformed into high-resolution (HR) ones by super-resolution technique, and then train the model on HR images; or a shared feature space is created for learning LR and HR images \cite{wang2016studying,zou2011very,Lu2018dcr}. However, even with the super-resolution process, details are predicted rather than real information, so classifier may not enhance the discriminative power from enlarged images.

There are also works that use knowledge distillation \cite{Distillation_2018Low, lr_kd1, tip2020-ge2020efficient,ddl2020,ge2020look}  to make an LR network mimic a HR network which is obtained under a rich HR training set. With paired LR and HR images of the same identity, the LR network is supervised by both class label and the soft target of HR network, corresponding to classification loss and distillation loss, respectively. The distillation process is to make student network output close to the soft target of teacher's output, narrowing the performance gap between the HR network and the LR network. But it is difficult for the existing knowledge distillation methods to control what and how to learn from the teacher, which requires a lot of tuning parameters, such as the temperature parameter and the balance coefficient between classification and distillation. 

\begin{figure}[h]
    \centering
    \includegraphics[width=0.8\textwidth]{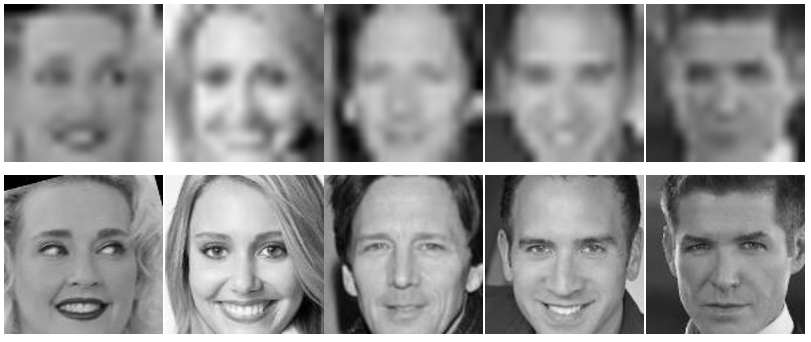}
    \caption{The first and second rows are LR and corresponding HR images in the training set CASIA, the resolutions are $16\times16$ and $120\times120$ respectively.}
    \label{fig:lr_hr}
\end{figure}

\begin{figure}[h]
    \centering
    \includegraphics[width=0.8\textwidth]{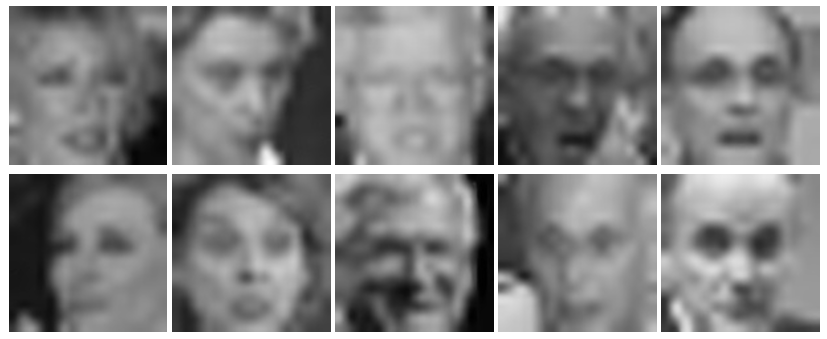}
    \caption{Positive pairs of every column in LR setting of LFW, where the resolution is $16\times16$.}
    \label{fig:positive}
\end{figure}

\begin{figure}[h]
    \centering
    \includegraphics[width=0.8\textwidth]{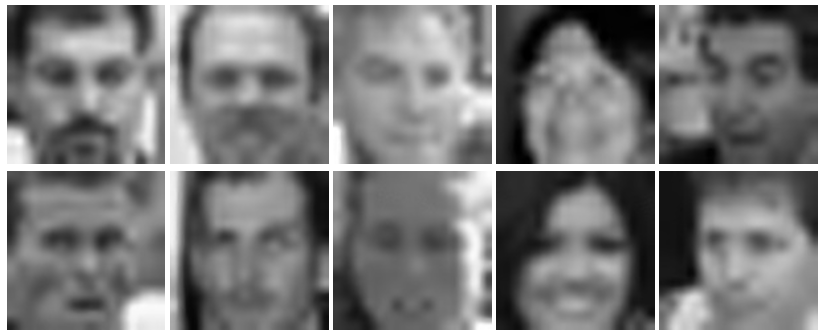}
    \caption{Negative pairs of every column in LR generation of LFW, where the resolution is $16\times16$.}
    \label{fig:negative}
\end{figure}

\begin{figure}[ht]
    \centering
    \includegraphics[width=0.9\textwidth]{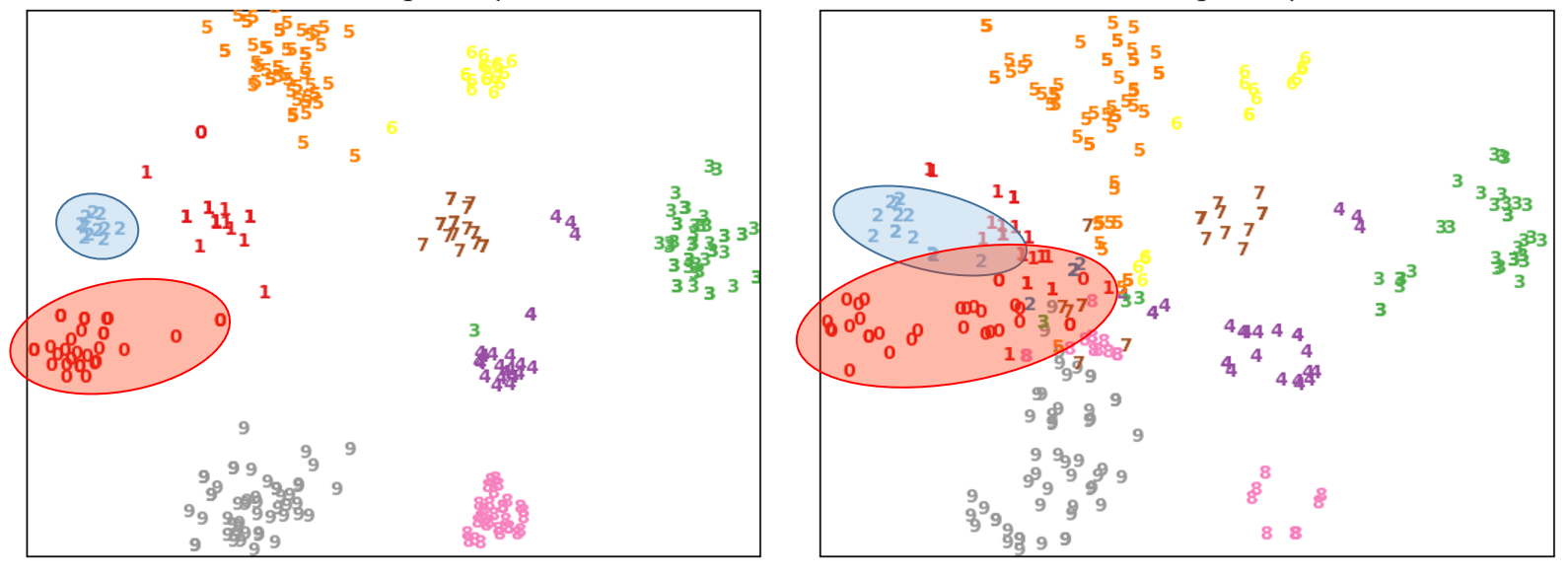}
    \caption{The left and right figures illustrate the hidden space distribution of features of HR and corresponding LR faces. The numbers represent different people. }
    \label{fig:confusion}
\end{figure}

As illustrated in Fig. \ref{fig:lr_hr}, there are obvious differences between LR and HR face images. LR face images lose a lot of details and lighting information of the face, and have only blurred outlines. In the testing phase of LR setting, it is required to determine whether the LR image pairs are from the same person. As demonstrated in Fig. \ref{fig:positive} and Fig. \ref{fig:negative}, it is difficult for even humans to tell whether these face pairs are from the same person or not. It is observed in Fig. \ref{fig:confusion} that the learned deep features of LR faces tend to have larger variation than those of HR faces\footnote{Download VGGFace2 \cite{cao2018vggface2} with SENet-50 from \url{https://github.com/cydonia999/VGGFace2-pytorch}  for feature extraction.}. As a result, the LR faces of similar persons are easily overlapped in the feature space, leading to high recognition error rate.


In this paper, we devise a new margin-based method for LR face recognition. The existing margin-based methods only consider margin penalization on the target class for each face image, and it still does not work well for the persons with embeddings close in the feature space. When competing for the correct label, the rival class that has the closest embedding to the target is the main cause of recognition error. This situation is much worse under LR setting as indicated in Fig.~\ref{fig:confusion}.
Motivated from the idea of Rival Penalized Competitive Learning (RPCL) \cite{rpcl_xu1993rival,xu2007scholarRPCL}, we further impose margin into the rival class to force the deep neural network to learn more discriminative features. RPCL is a further development of competitive learning by appropriately balancing a participating mechanism and a leaving mechanism. When the centers compete for each coming sample, not only the winner is learned towards the sample but also the rival (i.e., the second winner) is pushed away a little bit from the sample for compact and discriminative information allocation. Based on this property, we present a deep RPCL learning for face recognition. Our main contributions can be summarized from the following aspects:
\begin{itemize}
  \item We present an RPCL-based method to enhance the discriminative power of the deeply learned LR features. This is fulfilled by imposing a margin, for each input LR face image, into the rival logit which is defined as the highest non-target logit. The rival margin is devised as a reverse direction in contrast to the penalization on the target logit, and it enforces a de-learning on the CNN parameters, which would push the new embedding of the input sample away from the rival identity.
  
  \item Our method is not a trivial application of RPCL over the centers of each identity during training, because the embedding by the deep CNN changes as the network parameters change. Moreover, discriminative power conveyed by face centers is not able to generalize to open-set face verification. Instead, both the target margin and the rival margin are imposed to guide the parameter learning of CNN. The learned features are not only compact in the same class but also separate in competitive classes.
  
  \item Our deep RPCL can improve the existing margin-based methods or the data-uncertainty based methods, especially in low-resolution face recognition. This is demonstrated by comprehensive experiments on benchmark datasets, in comparisons with state-of-the-art methods.
\end{itemize}

\section{Related work} \label{sec:related work}

\subsection{Margin methods based on softmax classifier}

With the emergence of a large amount of face data, deep CNNs have proved their strength in face detection, face alignment, and facial recognition. DeepFace \cite{Deepface_taigman2014deepface} first used deep CNNs to extract facial features, and it laid the basic framework for face detection, alignment, cropping, CNN extraction, and classification. Later, DeepID \cite{DeepID_sun2014deep, DeepID2+_sun2015deeply, DeepID3_sun2015deepid3} and FaceNet \cite{facenet_Schroff2015FaceNet} made some improvements based on DeepFace, by increasing the data set, adding identification signals and verification signals, or by a triplet loss to enforce intra-class compactness and inter-class discrepancy. Softmax classifier is used in this line of methods to classify different identities in the training set.

The softmax loss has been found to have some drawbacks in generalizing the discriminative power from the training set to the open-set face verification problem. One popular way is to incorporate margins in the logit of the softmax. For example, Sphereface \cite{sphereface_liu2017} introduced a multiplicative angular margin penalty to the target logit, which corresponds to the identity of the input face image. CosFace \cite{cosface_wang2018} penalized the target logit directly by adding a cosine margin. ArcFace \cite{arcface_deng2019} advocated an additive angular margin loss to improve the discriminative power and to stabilize the training process. The above margin-based methods all reached a very high accuracy in high-quality images, e.g., $99\%+$ in LFW dataset \cite{LFW_Gary2008}. However, it is still far from perfect when applying them in low-resolution images, e.g., from the urban video surveillance system.


\subsection{Face embedding by triplet loss and others}

The face recognition model can be trained to directly learn discriminative features. FaceNet \cite{facenet_Schroff2015FaceNet} trained the deep CNNs using a triplet loss, i.e., two matching face features of the same identity and a non-matching face feature from a different identity form a triplet, and a Euclidean margin was added to enlarge the distance between the positive pair and the negative. In \cite{Wen2016centerloss}, a center loss was proposed to minimize the intra-class variance of the deep features, in combination with the softmax loss. In \cite{arcface_deng2019}, other loss functions, e.g., Intra-Loss and Inter-Loss, were also designed based on the angles between the face feature and the ground-truth center, and the angles between different centers, respectively.

The margin-penalized softmax loss, e.g., CosFace or ArcFace, are essentially related to triplet loss, if we assign the identity label to the image embedding by finding the nearest neighbor rather than by computing the closest distance or highest similarity to the center vector representing the face identity. Thus, the triplet loss is to penalize the target distance with a Euclidean margin based on the k-nearest neighbors (k-NN) classifier. Here, the goal of this paper is not to propose a new type of margin loss, but to enforce a rival penalized mechanism into the deep representation learning. Our deep RPCL method not only penalizes the target similarity but also oppositely penalizes the rival similarity for more discriminative power. As a result, we can develop deep RPCL versions of CosFace, ArcFace, triplet loss, and also center loss.


Probabilistic Face Embedding (PFE) \cite{PFE_shi2019probabilistic} is the first to consider uncertainty by modeling each face image embedding as a Gaussian distribution, with the mean being the pre-trained face feature and the variance capturing the uncertainty of the feature. Later, the work in \cite{DUL_chang2020data} proposed to learn the mean and the variance simultaneously by data uncertainty learning (DUL) with a KL-divergence constraining the Gaussian to be close to the standard normal distribution. Our method is compatible with PFE or DUL, i.e., they can be used jointly for face recognition.

\subsection{LR face recognition}

LR face recognition has attracted increasing studies over the past few years. Although the existing HR face recognition methods show outstanding performance, these models do not perform well when directly deployed on LR setting. The distributions between HR and LR faces are very different as demonstrated in Fig.~\ref{fig:lr_hr}-\ref{fig:confusion}. Since massive wild or surveillance LR faces are not available for training, the existing LR methods are based on down-sampling HR faces. \cite{lr_explore_aghdam2019exploring} investigated the face crop factors that would affect the LR face recognition performance, and trained the model by leveraging the factors to improve the accuracy. \cite{liPei2019low} conducted experimental research in actual surveillance applications, and empirically evaluated super-resolution methods for LR face recognition. Experiments showed that the recognition performance decreases as the resolution of the face image reduces. Basically, two stream of efforts have been made to advance LR face recognition.

The first is to embed HR and LR faces into the same feature space through a shared network \cite{ren2012coupled, biswas2011multidimensional, Lu2018dcr, Fine-to-coarse_peng2016fine}. Mixed-training method \cite{Fine-to-coarse_peng2016fine} has been demonstrated to be effective in LR face recognition when training on both HR and LR images, or training on HR images first and then fine-tuning on LR images. The DCR model \cite{Lu2018dcr} consisted of a big trunk network, to learn discriminative features shared by face images of various resolutions, and a branch network, to reduce the distance between the HR feature and its LR counterpart. \cite{wang2016studying} proposed a partially coupled network, and suggested that partial network sharing has more flexibility and better results than full sharing.

The second is knowledge distillation by considering HR network as teacher and LR network as student, respectively. Abundant knowledge is transferred from teacher to student, so that the student’s recognition ability is equivalent to that of the teacher. \cite{Distillation_2018Low} aimed to address the problem of identifying LR faces with extremely low computational cost via selective knowledge distillation. \cite{tip2020-ge2020efficient} proposed that the direct transfer from private HR to wild LR may be difficult, and used public HR and LR as a bridge to distill and compress knowledge. In order to fully transfer HR recognition knowledge to LR network, \cite{ge2020look} not only focuses on the first-order knowledge learning between points, but also considers high-order distillation. The knowledge of various order relationships is extracted from the teacher network as a supervision signal for student network. 

The above works devise and train LR network by utilizing the common characteristics between HR and LR faces. In this paper, we adapt the HR network to LR recognition by tackling the difficulty of LR induced large variation of the deeply learned features. This is fulfilled by further imposing a deep penalization on the rival class, to control the variation due to the increasing uncertainty when the resolution becomes low.

\section{Method} \label{sec:method}

\subsection{A brief review of RPCL}

\begin{figure}[ht]
    \centering
    \includegraphics[width=0.8\textwidth]{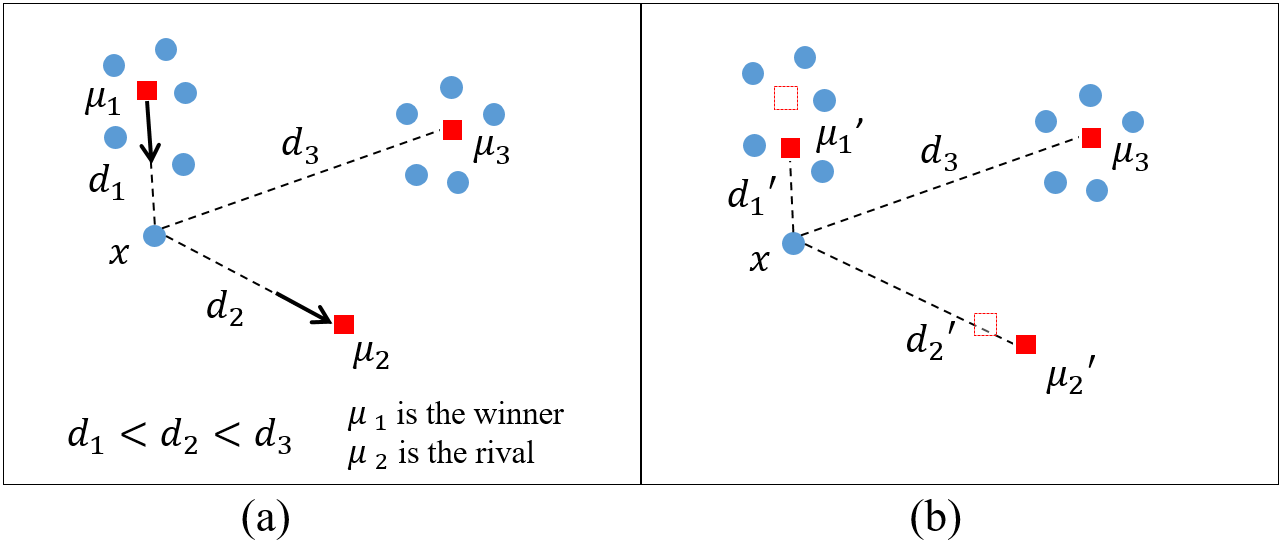}
    \caption{A sketch of the original RPCL learning. The figure is adapted from Figure 6 in \cite{xu2007scholarRPCL}. Red $\mu$ denotes class center, and $d$ represents the distance between the blue sample $x$ and the class center. The first and second class centers closest to the sample point are winner and rival, respectively. In (a), since $d_1<d_2<d_3$, $\mu_1$, $\mu_2$ are winner and rival. In (b), RPCL enforces the winner to approach sample and the rival to stay away from it. $\mu_2$ is a threat, a potential competitor for $\mu_1$.}
    \label{fig:rpcl_win_rival}
\end{figure}

Proposed in the early 1990s for clustering analysis \cite{rpcl_xu1993rival}, RPCL is a further development of competitive learning by appropriately balancing a participating mechanism and a learning mechanism such that an appropriate number of clusters will be determined \cite{xu2007scholarRPCL}. Specifically, as shown in Fig. \ref{fig:rpcl_win_rival}, suppose we use parametric centers $\{\mu_k\}_{k=1}^K$ to cluster the data $\{x_i\}_{i=1}^N$. For each coming data point $x_i$, not only the winner center is learned towards it but also its rival (i.e., the second winner) is pushed away a little bit from it to reduce a duplicated information allocation:
\begin{align}
    & \mu_k^{new} \leftarrow \mu_k^{old} + \eta p_{ki} (x_i-\mu_k^{old}), 
    \label{eq:rpcl} \\
    & p_{ki} =\left\{ \begin{array}{ll}
        1, & \text{ if } k=c, c=\arg\min_k d(x_i,\mu_k); \\
        -\gamma, & \text{ if } k=r, r=\arg\min_{k\neq c} d(x_i,\mu_k); \\
        0, & \text{ otherwise },
    \end{array} \right.  \nonumber
\end{align}
where $\gamma>0$ is a small number for controlling the penalizing strength, $d(x_i,\mu_k)$ measures the distance between $x_i$ and $\mu_k$, e.g., $\|x_i-\mu_k\|^2$.

\subsection{Deep RPCL for face recognition}

\begin{figure*}[ht]
    \centering
    \includegraphics[width=1.0\textwidth]{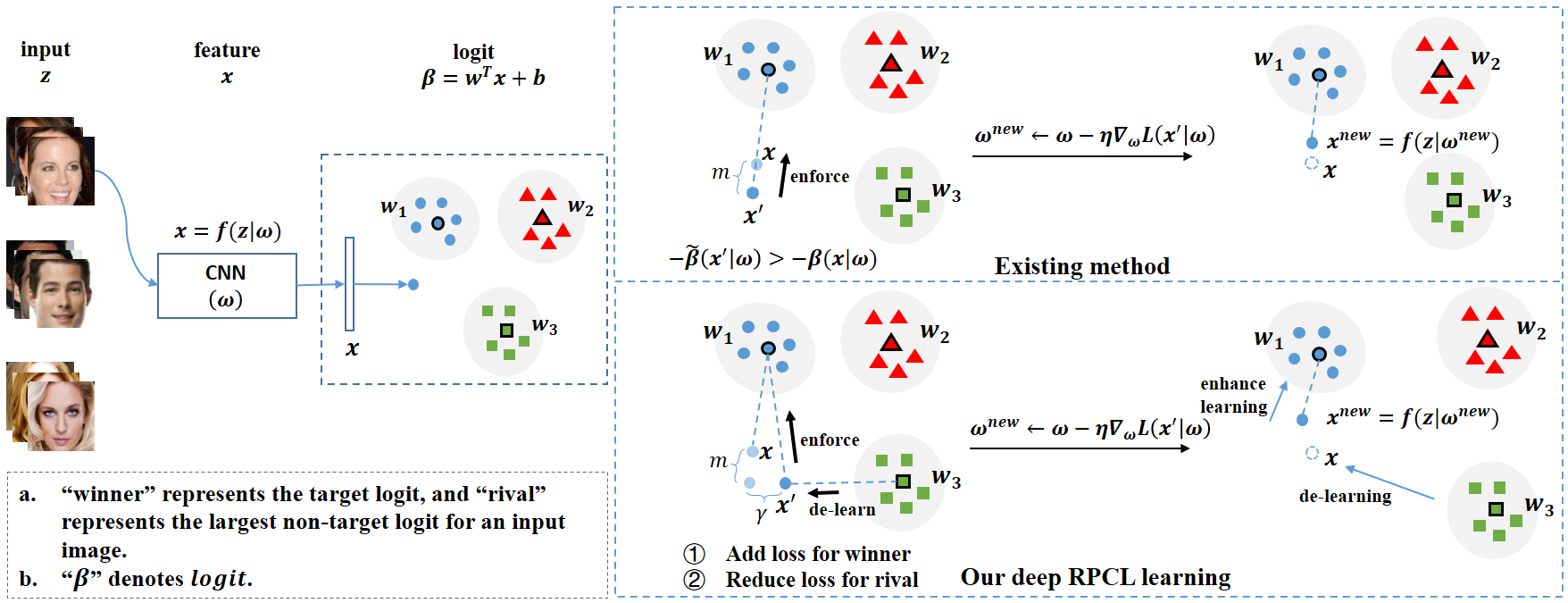}
    \caption{An overview of deep RPCL learning for face recognition in comparisons with existing methods. Neural network $f(z|w)$ map input image $z$ to face feature $x$ in latent space, and the classification logit $\beta$ are obtained through a fully connected network. $w_i$ denotes class center. Suppose $x$ belongs to class $w_1$ and $w_1$ is the target center. We compute the distance from the predictive probability. Existing method increases the distance between $x$ and target center $w_1$ on loss function, with $x$ becoming $x'$, and finally achieves the goal of enforcing intra-class compactness by punishing target logit. Our deep RPCL learning not only increases the distance between $x$ and target center $w_1$, but also takes into account the closest non-target center $w_3$ (rival) and shortens the distance between $x$ and $w_3$, with $x$ becoming $x'$. Finally, $x$ can get closer to the target center and away from potential threats. At the same time, class $w_3$ will have less interference and establish a clearer boundary.}
    \label{fig:RPCL-face}
\end{figure*}


As illustrated in Fig.~\ref{fig:confusion}, LR faces are blurry, lack personalized details, and different subjects overlap more easily than HR faces in the feature space, making it difficult for classification. RPCL enforces a regulation on the largest or closest non-target, which suggests that RPCL would increase the discriminative power in representation learning.  With the help of RPCL, similar but not the same subject would be pushed far away from each other in the feature space, which would make classification easier and more accurate. Thus, not applying RPCL directly in the raw face sample space, we develop a deep RPCL learning over the feature space for LR face recognition.

We start from the commonly-used softmax loss, which is
\begin{equation}
   L_{s}=-\frac{1}{N} \sum_{i=1}^{N} \log p_i
   =-\frac{1}{N} \sum_{i=1}^{N} \log \frac{e^{\beta_{y_i}}}{\sum_{j=1}^{n} e^{\beta_j}} ,
   \label{eq:softmax-loss}
\end{equation}
where $p_i$ denotes the posterior probability of the face feature $x_i$ being correctly classified to the target identity $y_i$, $n$ and $N$ is the number of classes and samples, and $\beta_j$ is the logit score of $x_i$ to the identity $j$, usually given by the activation of a fully-connected layer with weight vector $W_j$ and bias $b_j$. For simplicity, we may ignore the subscript of $x_i$ when there is no confusion. Then, we have
\begin{equation}
    \beta_j = W_j^Tx + b_j.
    \label{eq:logit-betaj}
\end{equation}
Following \cite{sphereface_liu2017, cosface_wang2018, arcface_deng2019}, we fix $b_j=0$, and set $\|W_j\|=1$, $\|x\|=s>0$. It follows from $W_j^Tx=\|W_j\|\cdot\|x\|\cos{\theta_j}=s\cos{\theta_j}$ that the logit $\beta_j$ can be represented as a function of $\theta_j$ in the angle space, where $\theta_j$ is the angle between the weight vector $W_j$ and the feature $x$.

The weight vector $W_j$ can be regarded as the center of the face features corresponding to the identity $j$. Thus, the logit by Eq.(\ref{eq:logit-betaj}) measures the similarity between the face feature $x$ and the center $W_j$. It can be further noted that the logit is equivalent to the negative Euclidean distance because $-d(x,W_j)=-\|x-W_j\|^2=-\|W_j\|^2-\|x\|^2+2W_j^Tx = 2\beta_j -1-s^2$. As a result, one direct application of RPCL for deep face recognition is to use $W_j$ as cluster centers. The difference from Eq.(\ref{eq:rpcl}) is that the winner is given by the target identity because we would like $W_{y_i}$ to get close to the feature $x_i$ for smaller clustering error. The rival is defined as the non-target identity which has the highest logit value. Ideally, enforcing RPCL learning over the face feature centers will further separate the centers from each other. In real practice, if the cluster center is disturbed or moved inappropriately, the neural network may be disordered and difficult to train and converge. Moreover, it is also difficult to generalize the trained centers to the open-set face verification problem, which requires learning discriminative features for the input face images of identities different from the training set. Therefore, we need an RPCL scheme that can affect the parameter learning over deep CNN.

Suppose the face feature $x$ is the embedding of the input image $z$ by a deep CNN $f(\cdot|\omega)$, i.e., $x=f(z|\omega)$ with $\omega$ being the network parameters. Notice that $x$ changes as $\omega$ updates. Thus, learning over the face features is different from the traditional RPCL which is conducted over fixed observations. Here, inspired by Eq.(\ref{eq:rpcl}) and the existing margin loss methods \cite{sphereface_liu2017,cosface_wang2018,arcface_deng2019}, we develop an RPCL-based margin loss for deep face recognition as follows:
\begin{align}
    & L_{r p c l}=-\frac{1}{N} \sum_{i=1}^{N} \log \frac{e^{\beta_{y_{i}}}}{\sum_{j=1}^{n} e^{\beta_{j}}}, 
    \label{eq:loss-rpcl} \\
    &  \beta_{j}=\left\{
            \begin{array}{cl}
            g\left(x_{i}, W_{j}, m\right), & \text { if } j=y_{i}; \\
            g\left(x_{i}, W_{j},-\gamma\right), & \text { if } j=\arg \max _{k \neq y_{i}} \beta_{k}; \\
            \beta_{j}, & \text { otherwise; }
            \end{array}\right.
    \label{eq:betaj-rpcl}
\end{align}

where $g(x_i,W_j,m)$ is designed as a function to compute the logit, satisfying that $g(x_i,W_j,m)=\beta_j$ if the margin $m=0$; $g(x_i,W_j,m)<\beta_j$ for some given $m>0$; and $g(x_i,W_j,-\gamma)>\beta_j$ for some given $\gamma>0$, $\gamma \ll m$. For example, if the logit is computed by Eq.(\ref{eq:logit-betaj}) with an additive margin, then $g(x_i,W_j,m)=W_j^Tx -m$.

In the RPCL-based margin loss by Eq.(\ref{eq:loss-rpcl})\&(\ref{eq:betaj-rpcl}), we have introduced a margin $m$ to the winner (or the target logit), and it plays the same role as in the existing margin methods, e.g., CosFace \cite{cosface_wang2018} or ArcFace \cite{arcface_deng2019}. The margin $m$, which reduces the logit to be lower, leads to a more stringent loss for classification and induces an extra margin away from the original classification boundary. This configuration updates the network parameters to compute the new face features that are more discriminative. However, the existing margin method only has the winner part in Eq.(\ref{eq:loss-rpcl})\&(\ref{eq:betaj-rpcl}), by paying all the attention to the probability of the target winning. They ignore the non-target class that has the largest probability value, which is potentially the biggest threat to the classification. 

Different from the existing margin methods, we impose a margin $\gamma$ to the rival (or the largest non-target logit), as in Fig.\ref{fig:RPCL-face}. The rival identity is the main competitor of the face image, i.e., the chief contributor to the classification error. Thus, we directly discourage the parameter learning by a margin in an opposite direction to the target logit, so that the rival logit is pushed down a bit, while still keeping the non-target non-rival logits unchanged. As a result, the deeply learned face features become more discriminative than the existing methods which only penalize the target logit.

When $\gamma=0$ in Eq.(\ref{eq:betaj-rpcl}), the reverse margin for the rival vanishes, and then the RPCL-based loss degenerates back to the existing margin methods, like CosFace or ArcFace. Moreover, in the scenario of two classes, the rival is always the other non-target identity. Taking $g(x,W_j,m)=W_j^Tx-m$ for example, the class decision boundary becomes $W_1^Tx-m = W_2^Tx+\gamma$, or $W_1^Tx-(m+\gamma) = W_2^Tx$. This scenario is equivalent to penalize the target logit with a larger margin $m+\gamma >m$. In the cases of $n>2$ classes, the rival varies from any non-target identities, and the reverse margin $\gamma$ provides a unique force to derive intra-class compactness and inter-class discrepancy. In real applications, $n$ is usually very large, up to thousands.

\subsection{Deep RPCL for CosFace, ArcFace, DCR}

According to Eq.(\ref{eq:loss-rpcl})\&(\ref{eq:betaj-rpcl}), our RPCL-based margin loss can be developed by considering different forms of the logit computing function $g(x,W_j,m)$. In the following, we provide the corresponding RPCL versions of CosFace and ArcFace, by considering margin penalization in the cosine space or angle space.
Suppose we have set $\|W_j\|=1$ and $\|x\|=s>0$. For CosFace, we have
\begin{align}
    g_c(x,W_j,m) = W_j^Tx-sm = s(cos\theta_j -m),
    \label{eq:rpcl-cos-g}
\end{align}
where $\theta_j$ is the angle between the weight vector $W_j$ and the face feature $x$. For ArcFace, the margin $m$ is directly considered in the angle space, i.e.,
\begin{align}
    g_a(x,W_j,m) = s(cos(\theta_j +m)).
    \label{eq:rpcl-arc-g}
\end{align}
Notice that both $g_c$ and $g_a$ by Eq.(\ref{eq:rpcl-cos-g})\&(\ref{eq:rpcl-arc-g}) satisfy $g(x,W_j,m)<g(x,W_j,0)$ for certain $m>0$, i.e., the margin $m$ enforces a more strict level for the logit. Otherwise, $g(x,W_j,-\gamma)>g(x,W_j,0)$ for certain $\gamma>0$, enforcing a loose level for parameter de-learning.

Different from CosFace and ArcFace, center loss \cite{Wen2016centerloss} is to directly minimize the distances between the deep features and their corresponding class centers, and is used as a regularization term for the softmax loss. Center loss has been used in DCR \cite{Lu2018dcr} to demonstrate robustness in LR face recognition. So, we also develop DCR into an RPCL version for experimental comparisons. Specifically, we add a rival loss to the DCR loss as follows:
\begin{align} \label{eq:dcr_rpcl}
  &L_d=L_{s} + \beta \Sigma_{i=1}^{n}\left\|x_{i}-{c}_{y_{i}}\right\|_{2}^{2}-\gamma \Sigma_{i=1}^{n}\left\|{x}_{i}-{c}_{r}\right\|_{2}^{2},
\end{align}
where $L_s$ is the softmax loss given by Eq.(\ref{eq:softmax-loss}), $\beta>0$ controls the strength of the center loss for the winner center (i.e., the target center), and $-\gamma<0$ enforces a reverse penalization pushing $x_i$ away from the rival center.

\newcommand{\tabincell}[2]{\begin{tabular}{@{}#1@{}}#2\end{tabular}}  

\section{Experiments} 


\subsection{Settings} \label{subsec:settings}

\textbf{Training datasets.} 
There are many general face recognition training sets, such as CASIA-WebFace \cite{casia_yi2014learning} and MS-Celeb-1M \cite{celeb_guo2016ms}. We choose CASIA-WebFace as the training set, and it contains 10,575 subjects, a total of about 500,000 face images crawled from the web. For data preprocessing, we adopt insightface\footnote{https://github.com/deepinsight/insightface} to do face detection and face alignment, and finally crop to the target size ($120 \times 120$) as HR images. Then we use the public code\footnote{https://github.com/yoon28/unpaired\_face\_sr} of a recent specialized LR face generation model \cite{bulat2018learn} to generate LR samples ($16 \times 16$) to simulate real-world LR face images.

\textbf{Benchmarks.} 
We use two LR benchmark datasets, SCFace \cite{scface_grgic2011scface} and TinyFace \cite{cheng2018tinyface}, to evaluate the LR recognition performance. Moreover, we use state-of-the-art LR face generation model\cite{bulat2018learn} to generate LR samples set that simulate real-world LR face images for more LR testing datasets from LFW \cite{LFW_Gary2008}, YTF \cite{YTF_Wolf2011Face}, MegaFace \cite{megaface_kemelmacher2016megaface}. The corresponding generated LR versions are denoted as LFW$^*$, YTF$^*$, MegaFace$^*$.

\textbf{Evaluation metrics.} 
We consider both face verification and face identification in testing. Face verification is to judge whether the given pair of images are from the same person by computing cosine distance according to the two face feature vectors. Face identification is to find the image most similar to the target face from the gallery set through cosine distance comparison.

\subsection{Training details} \label{sec:training-details}

Recent state-of-the-art face recognition methods, ArcFace \cite{arcface_deng2019}, CosFace \cite{cosface_wang2018} and SphereFace \cite{sphereface_liu2017}, are included for experimental comparisons. They are all implemented under the PyTorch framework, on the basis of the ArcFace project\footnote{https://github.com/ronghuaiyang/arcface-pytorch}. The RPCL learning for CosFace and ArcFace are developed by taking Eq.(\ref{eq:rpcl-cos-g})\&(\ref{eq:rpcl-arc-g}) into the Eq.(\ref{eq:loss-rpcl})\&(\ref{eq:betaj-rpcl}). We set the initial learning rate to $0.1$, which is decremented by 10 times every 10 epochs, for a total of 50 epochs. We use Adam optimizer to find the optimal network parameters. Each training roughly takes 20 hours with two Titan Xp GPUs.



\subsection{On native unconstrained LR images}

In real surveillance applications, LR faces usually are collected under uncontrolled conditions in background, illumination, distance and cameras. In addition, usually the images collected in the gallery set are HR images, and the images in the probe set we need to verify are LR, blurry. The resolution levels are not matched between HR and LR images, and this setting requires the model's feature learning to be robust against varied resolutions. The SCFace \cite{scface_grgic2011scface} and TinyFace \cite{cheng2018tinyface} are two widely-used, real LR testing datasets.



\begin{table}[htbp]
    \centering
    \renewcommand\tabcolsep{9pt} 
    \begin{tabular}{l|lllll} 
        \toprule
        \diagbox{Method}{Distance}            & $1m$  & $2.6m$  &  $4.2m$   \\  
        \hline
        MDS \cite{MDS_mudunuri2015low}        & 69.5   & 66.0 & 60.3  \\
        DMDS \cite{DMDS_LDMDS_yang2017}       & 62.9   & 67.2 & 61.5  \\
        LDMDS \cite{DMDS_LDMDS_yang2017}      & 65.5   & 70.7 & 62.7  \\
        RICNN  \cite{RICNN_zeng2016}          & 74.0   & 66.0 & 23.0  \\
        LightCNN \cite{LightCNN_wu2018light}   & 93.8   & 79.0 & 35.8 \\
        VGGFace \cite{VGGFace_parkhi2015deep}  & 88.8   & 75.5 & 41.3 \\
        PeiLi's \cite{liPei2019low}            & 31.7 & 20.8 & 20.4  \\
        ResNet \cite{Wen2016centerloss}        & 94.3 & 81.1 & 36.3  \\
        DDL(ResNet50) \cite{ddl2020}   & 98.3 & \textbf{98.3} & 86.8 \\ 
        TCN \cite{tcn2019-zha2019tcn}  & \textbf{98.6} & 94.9 & 74.6 \\
        DCR \cite{Lu2018dcr}                  & 98.0  & 93.5 & 73.3   \\
        RPCL-DCR                              & \textbf{98.0}   & \textbf{98.0}  & \textbf{90.4} \\
        \bottomrule
    \end{tabular}
    \caption{Comparative results on the face images taken at three different distances from the SCFace test set. The longer the distance, the more blurred the face image.}
    \label{tab:scface}
\end{table}

\textbf{Results on SCFace.} SCFace \cite{scface_grgic2011scface} contains $4,160$ static face images of $130$ subjects. Images from different quality cameras and different distances simulate real-world situations. For each subject, there are one HR face image taken by the high-quality camera and $15$ LR face images that are taken by low-quality cameras at three different distances, $1 m \,(d3)$, $2.6 m\,(d2)$ and $4.2 m \,(d1)$. Following \cite{Lu2018dcr} \cite{liPei2019low} and \cite{discriminative_yang2017}, we randomly divide the data set into 80 individuals and 50 individuals, separately as the training set and the test set, and they do not intersect. HR images are used as gallery, LR images are used as probe. According to the cosine distance metric, we find the most similar face image from the gallery to match the face image in the probe, and calculate the rank-1 accuracy rate. To test the effectiveness of RPCL, we use DCR\cite{Lu2018dcr} as a baseline and develop it into RPCL versions by setting $\beta =0.008$, $\gamma =0.002$ in Eq.(\ref{eq:dcr_rpcl}). We also include other methods for comparisons, such as MDS \cite{MDS_mudunuri2015low}, DMDS \cite{DMDS_LDMDS_yang2017},  LDMDS \cite{DMDS_LDMDS_yang2017}, RICNN  \cite{RICNN_zeng2016}, LightCNN \cite{LightCNN_wu2018light}, and so on.

The DCR model is optimized with softmax loss and center loss, and it tackles the cross-resolution face recognition problem using training samples in different resolutions. We added RPCL to the center loss to impose a certain penalty on the rival by Eq.(\ref{eq:dcr_rpcl}), and found that it significantly improves DCR for the challenging distances at $4.2m$ and $2.6m$ as in Tab. \ref{tab:scface}. In particular, it increases the accuracy from $73.3.0\%$ to $90.4\%$ with an increment $17.1\%$ at $4.2m$, which is the most difficult scenario for all the methods.

It is observed that RPCL is more effective in dealing with LR, blurred images than HR ones. The LR images lose their personalized features, are much more ambiguous and difficult to distinguish than HR images. Then, in classification process, LR images have more potential threats that are very likely to be misjudged. RPCL takes into account the potential threats, while the existing face recognition methods only make the winner closer to the target, and ignore the potential threat (the second winner). Kicking out the threats is effective to enhance the discrminative power which is very important for classification problems.

\begin{table}[htbp]
    \centering
    \renewcommand\tabcolsep{4.5pt} 
    \begin{tabular}{l|c|llllll} 
        \toprule
        Methods      & RPCL   & Rank-1  & Rank-10  &  Rank-20   \\  
        \hline
        DCR \cite{Lu2018dcr}           & ×     & 0.29  & 0.40  & 0.44  \\
        RPCL\_DCR    & \checkmark   &  0.31 & 0.41  & 0.44 \\
        \hline
       PeiLi’s \cite{lipei2018low-survey} & ×   & 0.31 & 0.43  & 0.46  \\
        RPCL\_Lipei’s    & \checkmark    & 0.32 & 0.44  & 0.47 \\
        \hline
        CenterFace \cite{Wen2016centerloss}  & ×    & 0.32 & - & 0.45  \\
        RPCL\_CenterFace    & \checkmark      & 0.33 & 0.44  & 0.47 \\
        \hline
        Arcface \cite{arcface_deng2019}      & ×   & 0.26  & 0.34   & 0.37   \\
        RPCL\_ArcFace      & \checkmark        & 0.35  & 0.45   & 0.48   \\
        \hline
        CosFace \cite{cosface_wang2018}      & ×         & 0.29  & 0.39   & 0.42   \\
        RPCL\_CosFace       & \checkmark      & 0.32  & 0.41   & 0.44   \\
        
        \bottomrule
    \end{tabular}
    \caption{Face identification rank-1, rank-10 and rank-20 accuracy of different low-resolution models on TinyFace. }
    \label{tab:tinyface}
\end{table}

\textbf{Results on TinyFace.} TinyFace \cite{cheng2018tinyface} dataset provides a large scale native LR face images and contains large gallery population size. It was collected from snapshots of videos taken under surveillance conditions from a certain distance. We use the training set of TinyFace to fine-tune the basic model before 1:N identification test on the testing set of TinyFace. The results are given in Table \ref{tab:tinyface}.

We can observe that deep RPCL is able to improve the LR recognition methods, DCR and PeiLi's method, slightly but consistently under various scenarios. It should be noted that ArcFace and CosFace, which are originally devised for HR face recognition, perform poorly in LR face recognition, but they are improved by deep RPCL to outperform or at least comparable to DCR and PeiLi's method.

\subsection{On synthesized real-world LR images} \label{sec:comparison_lr}

\begin{figure}[H]
    \centering
    \includegraphics[width=0.7\textwidth,height=0.6\textwidth]{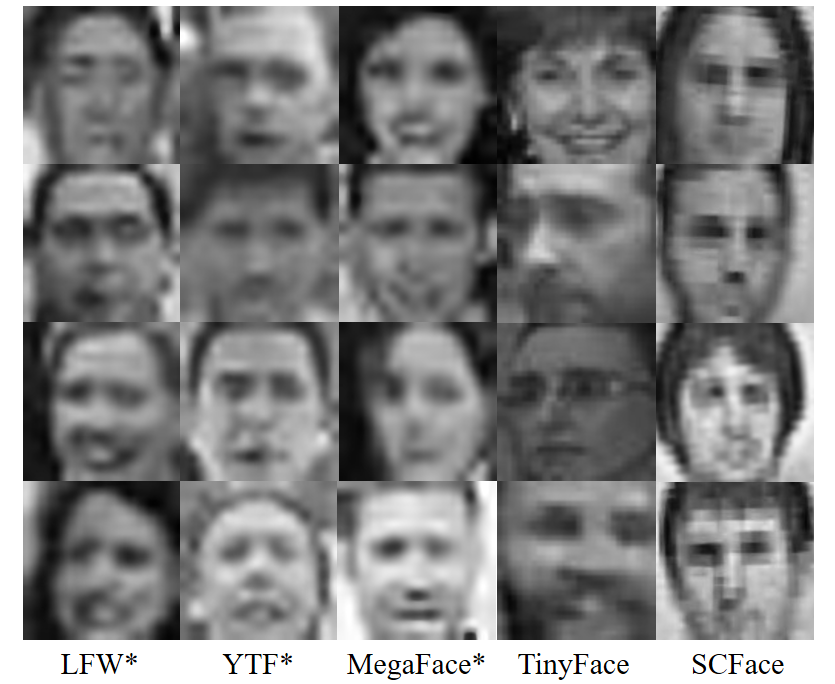}
    \caption{From the first column to the fifth column are: LFW*, TYF*, MegaFace*, TinyFace and SCFace. The first four columns are the low-resolution data set that we simulated close to the real-world. The last two columns are low-resolution images taken by the surveillance cameras.}
    \label{fig:lr_sample}
\end{figure}

To further evaluate the robustness of deep RPCL in LR face recognition, we conduct experiments on LR generations, i.e. LFW*, YTF*, and MegaFace*. We can check that the generated LR faces in the Fig. \ref{fig:lr_sample} are very close to the faces taken by the surveillance cameras.
We use $16\times 16$ as the LR image size. All the $16\times 16$ LR images are first enlarged to $120\times 120$ by a commonly-used super-resolution technology, VDSR\footnote{https://github.com/twtygqyy/pytorch-vdsr}. For fair comparisons, we build the same deep CNN to take $120\times 120$ images as input but train the network under the loss function of each method. 
For each setting of ArcFace by Eq.(\ref{eq:rpcl-arc-g}) or CosFace by Eq.(\ref{eq:rpcl-cos-g}), using SE-Block or not, we implement the RPCL scheme (by setting $\gamma>0$ in Eq.(\ref{eq:betaj-rpcl})) or not (by setting $\gamma=0$). 
Regarding the DUL \cite{DUL_chang2020data} experiment, we asked the original authors for some key codes for the experiment. To the best knowledge from our practical experience, we set the coefficient of KL loss as $0.01$.



The results are summarized in Table \ref{tab:compare_lr}. Again, we observed that deep RPCL is able to enhance the discriminative power of learned features, which is consistent with the results on real-world datasets.

\begin{table}[H]
    \centering
    \renewcommand\tabcolsep{4.5pt} 
    \begin{tabular}{l|l|cl|cl} 
        \toprule
        \multirow{2}*{Loss} & \multirow{2}*{Backbone} & \multicolumn{2}{c|}{LFW*} & \multicolumn{2}{c}{YTF*} \\ 
        ~ & ~ & No RPCL & RPCL & No RPCL & RPCL\\
        \midrule
        ArcFace & ResNet18 & 83.68 & \textbf{84.02} & 69.90 & \textbf{69.98}\\ 
        
        CosFace & ResNet18 & 83.45  & \textbf{83.48} & 68.64 & \textbf{68.66}\\
        
        \rowcolor{mygray}
        ArcFace & \tabincell{l}{ResNet18\\SE-Block} & 84.08 & \textbf{84.93} & 75.80 & \textbf{78.58}\\
        
        CosFace & \tabincell{l}{ResNet18\\SE-Block} & 82.97 &  \textbf{83.37} & \textbf{77.36} & 77.26 \\
        
        \rowcolor{mygray}
        \tabincell{l}{CosFace\\+DUL}  & \tabincell{l}{ResNet18\\SE-Block} & 83.92  & \textbf{84.70} & 78.14 & \textbf{78.80}\\
        \bottomrule
    \end{tabular}
    \caption{LR face recognition results on generated LR test sets. LFW*: the face image pair verification accuracy ($3,000$ positive face pairs, $3,000$ negative face pairs), YTF*: the face video pair verification accuracy. For each test set, we conducted a comparative experiment with or without PRCL. The last row refers to RPCL-based method on the basis of DUL network structure ($\mu$ and $\sigma$ branches).}
    \label{tab:compare_lr}
\end{table}

\begin{table}[htbp]
    \centering
    \renewcommand\tabcolsep{8.5pt} 
    \begin{tabular}{l|l} 
        \hline
        Method & Accuracy \\
        \hline
        SKD \cite{Distillation_2018Low} & 85.87 \\
        \hline
        Ge's TIP \cite{tip2020-ge2020efficient} & 85.88 \\
        \hline
        Ge's MM \cite{mm2019-ge2019fewer} & 93.96\\
        \hline
        ArcFace & 92.30 \\
         \hline
        CosFace  & 93.80 \\
         \hline
        RPCL-Arc & 94.70\\
         \hline
        RPCL-Cos & \textbf{95.13}\\
         \hline
    \end{tabular}
    \caption{Comparative experiment on LFW via bicubic down-sampling. The backbone of the last four methods is ResNet18.}
    \label{tab:lfw}
\end{table}

We also notice that some existing LR methods \cite{Distillation_2018Low, tip2020-ge2020efficient, mm2019-ge2019fewer} have been evaluated on LFW using bicubic to down-sample to $16\times 16$ and then up-sample to target size.
To make a fair comparison, we conduct a comparison under the same setting. The results in Table \ref{tab:lfw} again indicate the effectiveness of the deep RPCL to improve the LR face recognition. The original ArcFace and CosFace are not so good as Ge's MM \cite{mm2019-ge2019fewer} for this task, but surpass it when deep RPCL is activated.

\subsection{Empirical analysis on the intra-class compactness and inter-class discrepancy}

\begin{figure}[htbp]
    \centering
    \includegraphics[width=0.7\textwidth]{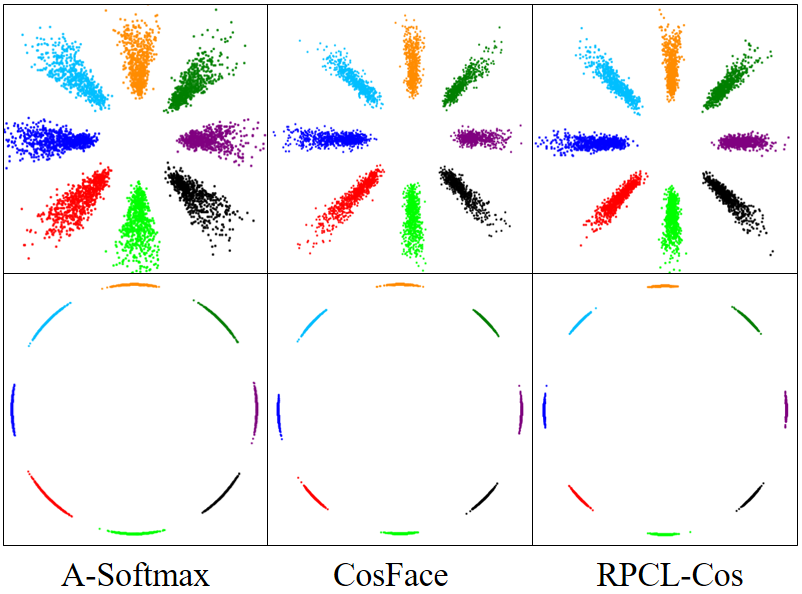}
    \caption{Visualization of 2-D face features from 8 people. The 1st row: shown in the Euclidean space. The 2nd row: mapped to the hypersphere. The columns from left to right: A-Softmax, CosFace, RPCL-Cos.}
    \label{fig:toy-example}
\end{figure}

For clear visualization, we select 8 people from the training set, and each person has more than 500 images. We train a network that uses 2-D feature to represent LR face ($16\times16$), and visualize the 2-D features in scatter plots with different colors for different persons. 
As shown in Fig. \ref{fig:toy-example}, RPCL-Cos demonstrates a larger classification margin between the classes than A-Softmax and CosFace. Moreover, we calculate the angle statistics in the Table \ref{tab:theta}. The results indicate that the face features are closer to their class centers, i.e., more intra-class compact, by RPCL-based methods when comparing to the corresponding CosFace or ArcFace baselines. This is because the punishment for rival keeps the face feature away from the class center that is most likely to be misjudged, reducing the possibility of misjudgment, and maximizing the class separability.

\begin{table}[H]
    \centering
    \renewcommand\tabcolsep{5pt} 
    \begin{tabular}{l|l|llll} 
        \toprule
          
          
              loss   & backbone & intra & inter  & w-w    & w-c \\
         \hline
          Softmax  & ResNet18 & 51.273 & 55.479 & 68.695 & 29.872  \\
          
         \hline
          
          ArcFace      & ResNet18 & 46.305 & 54.176 & 64.811 & 20.038  \\
          \hline
          \rowcolor{mygray}
          RPCL-Arc & ResNet18 & 45.631 & 54.271 & 71.098 & 19.316  \\
        \hline
          CosFace    & \tabincell{l}{ResNet18\\(SE-Block)} & 52.488 & 59.589 & 70.854 & 27.403   \\
          \rowcolor{mygray}
          RPCL-Cos  & \tabincell{l}{ResNet18\\(SE-Block)} & 49.331 & 59.645 & 69.632 & 24.283  \\
         \hline
          ArcFace    & \tabincell{l}{ResNet18\\(SE-Block)} & 55.315 & 59.421 & 72.763 & 29.944  \\
          \rowcolor{mygray}
          RPCL-Arc  & \tabincell{l}{ResNet18\\(SE-Block)} & 44.859 & 57.594 & 66.787 & 19.476 \\
          
        \bottomrule
    \end{tabular}
    \captionsetup{width=1.0\textwidth} 
    \caption{Statistics of angles $\in[0^{\degree}, 180^{\degree}]$ between two feature vectors. The "intra" refers to the mean of angles between each face feature vector and its corresponding class center. The "inter" refers to the mean of minimum angles between the class center and other centers. The "w-w" represents the mean of minimum angles between $W_{j}$'s. The "w-c" represents the mean angles between $W_j$ and its corresponding class center.}
    \label{tab:theta}
\end{table}

\begin{figure}[H]
    \centering
    \includegraphics[width=0.9\textwidth]{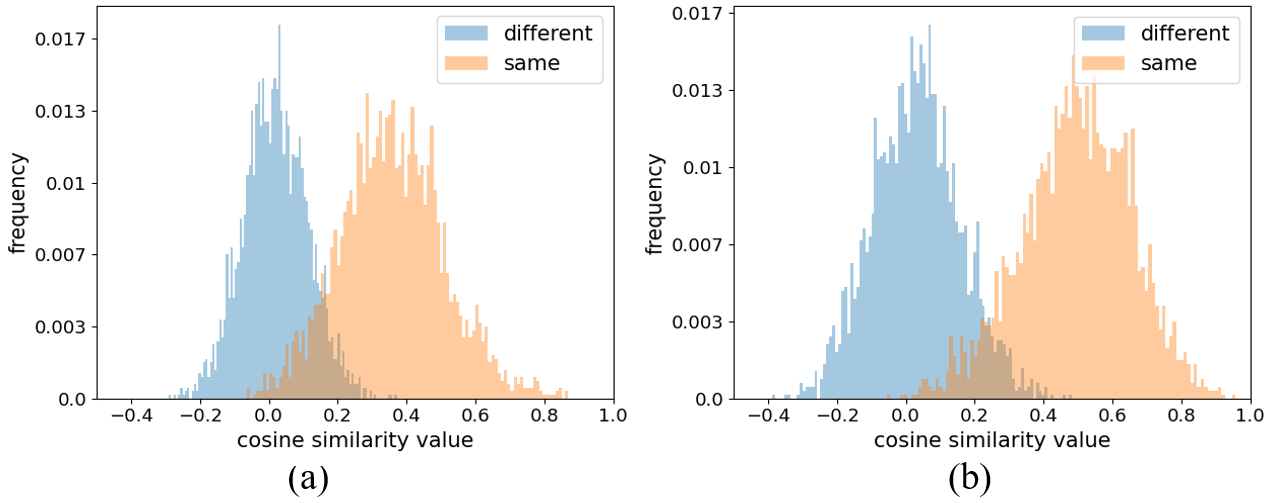}
    \caption{The distribution of cosine similarity score under the LR setting of LFW dataset by (a) ArcFace, and (b) RPCL-Arc.  The x-axis represents the cosine similarity of a pair of face images, while the y-axis is the frequency. Blue and orange respectively denote negative pairs (different persons) and positive pairs (the same person). }
    \label{fig:lfw}
\end{figure}

\noindent \textbf{Results on LFW.}
We calculate the distribution of the cosine similarity values between the positive pairs (the same person) and the negative pairs (different persons) under the LR setting of LFW dataset. As shown in Fig. \ref{fig:lfw}, the deeply learned face features are more discriminative and less overlapped by RPCL-Arc than by ArcFace, indicating that the RPCL's de-learning is effective in enhancing the discriminative power. In terms of Fisher discriminant criterion, i.e., the difference between the means of the two groups normalized by the within-group scatter, the computed Fisher criterion value for RPCL-Arc is $5.264$, much higher than $3.646$ for ArcFace. 

\begin{figure}[htbp]
\centering
\subfigure[CMC]{
\begin{minipage}[t]{0.5\linewidth}
\centering
\includegraphics[height=1.6in, width=2.3in]{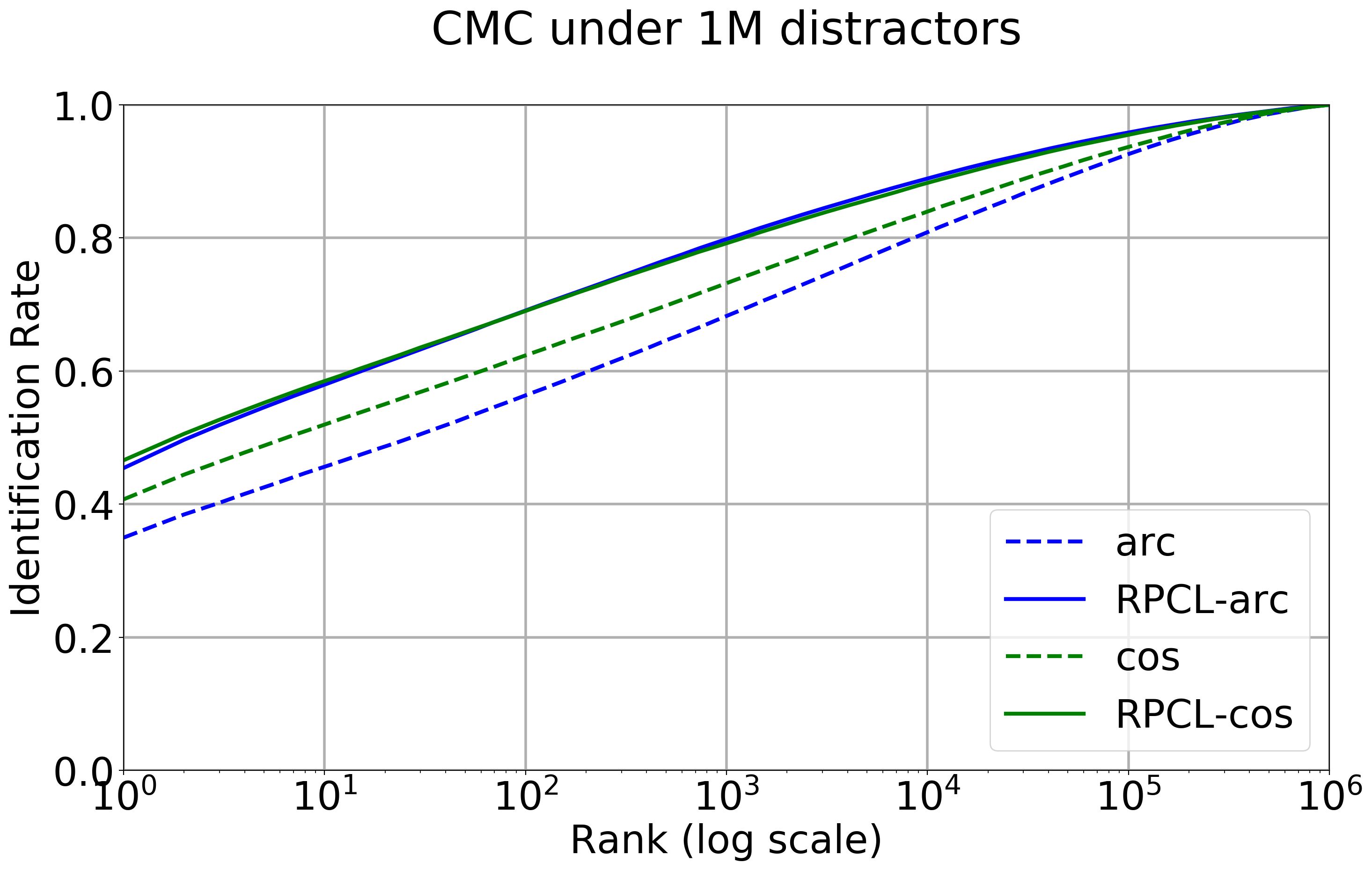}
\end{minipage}%
}%
\subfigure[ROC]{
\begin{minipage}[t]{0.5\linewidth}
\centering
\includegraphics[height=1.6in, width=2.3in]{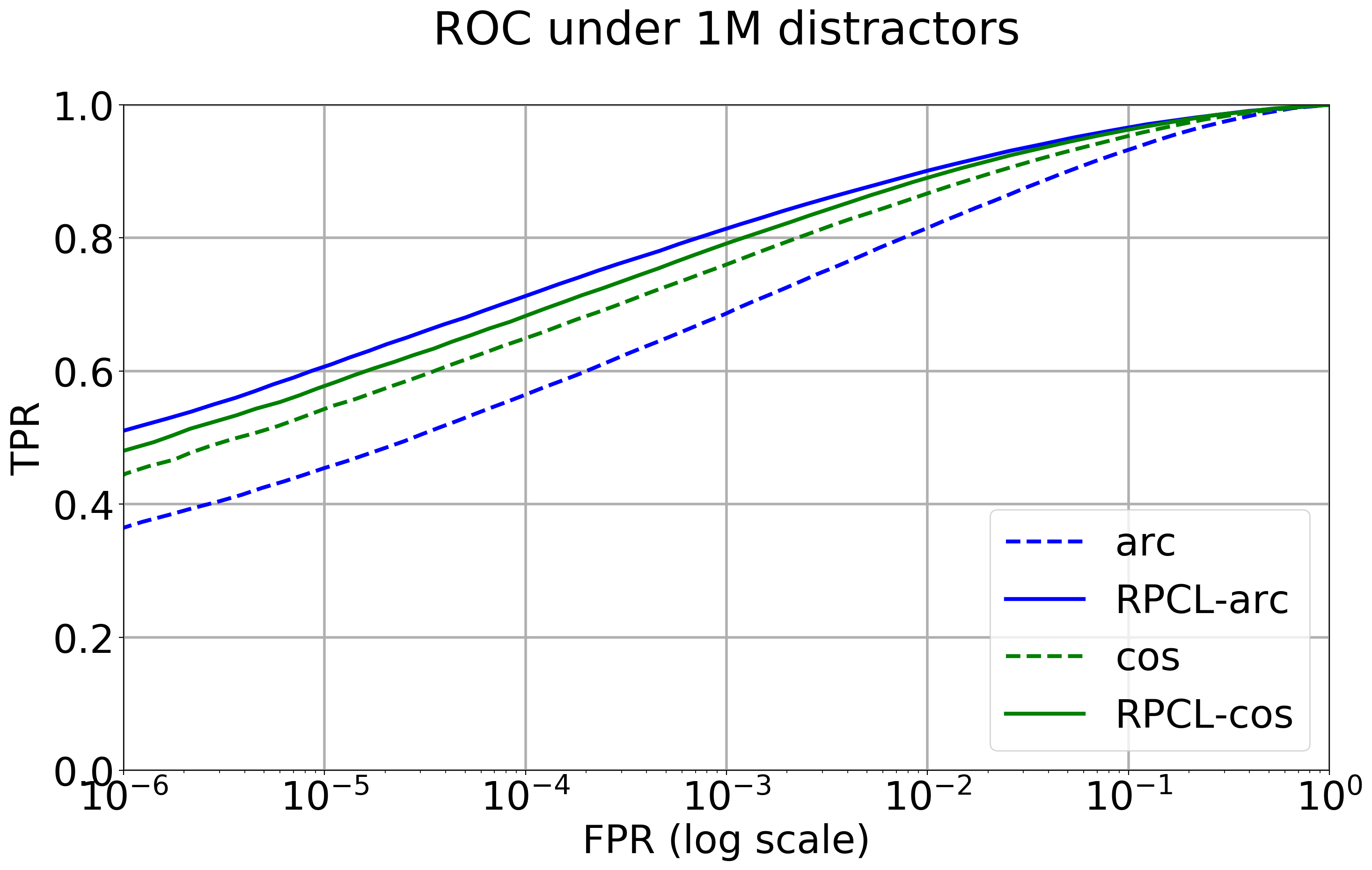}
\end{minipage}%
}%
\centering
\caption{CMC and ROC curves on MegaFace test set with 1M distractors. The dotted lines represent CosFace and ArcFace, and the solid line represents RPCL-based CosFace and ArcFace.}
\label{fig:megaface}
\end{figure}

\noindent \textbf{Results on Megaface.}
We train models with different loss functions on CASIA dataset and results are evaluated on Megaface in Fig. \ref{fig:megaface}. CMC refers to the rank-$k$ identification rate as $k$ grows, and ROC plots True Positive Rate (TPR) versus False Positive Rate (FPR). We can observe that an obvious improvement by RPCL from the CosFace and ArcFace baselines in terms of both CMC and ROC. 

\section{Conclusion}  \label{sec:conclusion}

In this paper, we have presented a deep RPCL representation learning for LR face recognition, which is a challenging problem because the variation induced from decreasing resolution tends to make the deep face features of different persons overlap with each other. The existing margin-based methods only consider penalization on the target logit, while our method further imposes a reverse margin into the rival logit, which is defined as the highest non-target logit for the input face feature. The reverse margin discourages the face feature from approaching the rival label, providing a de-learning force to make the deeply learned representation more discriminative. As a result, the face recognition accuracy can be improved. To verify the effectiveness of our method, we have conducted comprehensive experiments including face recognition in real LR benchmark datasets, or in generated LR images that simulate the real-world LR images. The results demonstrate that our method is particularly robust in LR face recognition tasks, and has good potential applications in real-world environments when HR or high-quality images are difficult to capture, e.g., in video surveillance applications.

\section*{Acknowledgment}
This work was supported by The National Key Research and Development Program of China (2018AAA0100700) of the Ministry of Science and Technology of China, and Shanghai Municipal Science and Technology Major Project (2021SHZDZX0102). Shikui Tu and Lei Xu are corresponding authors.

\bibliography{lpy-nn}

\begin{thebibliography}{10}
\expandafter\ifx\csname url\endcsname\relax
  \def\url#1{\texttt{#1}}\fi
\expandafter\ifx\csname urlprefix\endcsname\relax\def\urlprefix{URL }\fi
\expandafter\ifx\csname href\endcsname\relax
  \def\href#1#2{#2} \def\path#1{#1}\fi

\bibitem{Turk1991-eigen}
M.~A. {Turk}, A.~P. {Pentland}, Face recognition using eigenfaces, in:
  Proceedings. 1991 IEEE Computer Society Conference on Computer Vision and
  Pattern Recognition, 1991, pp. 586--591.

\bibitem{Deepface_taigman2014deepface}
Y.~Taigman, M.~Yang, M.~Ranzato, L.~Wolf, Deepface: Closing the gap to
  human-level performance in face verification, in: Proceedings of the IEEE
  conference on computer vision and pattern recognition, 2014, pp. 1701--1708.

\bibitem{DeepID_sun2014deep}
Y.~Sun, X.~Wang, X.~Tang, Deep learning face representation from predicting
  10,000 classes, in: Proceedings of the IEEE conference on computer vision and
  pattern recognition, 2014, pp. 1891--1898.

\bibitem{DeepID2_sun2014deep}
Y.~Sun, Y.~Chen, X.~Wang, X.~Tang, Deep learning face representation by joint
  identification-verification, in: Advances in neural information processing
  systems, 2014, pp. 1988--1996.

\bibitem{facenet_Schroff2015FaceNet}
F.~Schroff, D.~Kalenichenko, J.~Philbin, Facenet: A unified embedding for face
  recognition and clustering.

\bibitem{sphereface_liu2017}
W.~Liu, Y.~Wen, Z.~Yu, M.~Li, B.~Raj, L.~Song, Sphereface: Deep hypersphere
  embedding for face recognition, in: Proceedings of the IEEE conference on
  computer vision and pattern recognition, 2017, pp. 212--220.

\bibitem{cosface_wang2018}
H.~Wang, Y.~Wang, Z.~Zhou, X.~Ji, D.~Gong, J.~Zhou, Z.~Li, W.~Liu, Cosface:
  Large margin cosine loss for deep face recognition, in: Proceedings of the
  IEEE Conference on Computer Vision and Pattern Recognition, 2018, pp.
  5265--5274.

\bibitem{arcface_deng2019}
J.~Deng, J.~Guo, N.~Xue, S.~Zafeiriou, Arcface: Additive angular margin loss
  for deep face recognition, in: Proceedings of the IEEE Conference on Computer
  Vision and Pattern Recognition, 2019, pp. 4690--4699.

\bibitem{PFE_shi2019probabilistic}
Y.~Shi, A.~K. Jain, Probabilistic face embeddings, in: Proceedings of the IEEE
  International Conference on Computer Vision, 2019, pp. 6902--6911.

\bibitem{DUL_chang2020data}
J.~Chang, Z.~Lan, C.~Cheng, Y.~Wei, Data uncertainty learning in face
  recognition, in: Proceedings of the IEEE/CVF Conference on Computer Vision
  and Pattern Recognition, 2020, pp. 5710--5719.

\bibitem{Wen2016centerloss}
Y.~Wen, K.~Zhang, Z.~Li, Y.~Qiao, A discriminative feature learning approach
  for deep face recognition, in: B.~Leibe, J.~Matas, N.~Sebe, M.~Welling
  (Eds.), Computer Vision -- ECCV 2016, Springer International Publishing,
  Cham, 2016, pp. 499--515.

\bibitem{lipei2018low-survey}
P.~Li, P.~J. Flynn, L.~Prieto, D.~Mery,
  \href{http://arxiv.org/abs/1805.11519}{Face recognition in low quality
  images: {A} survey}, CoRR abs/1805.11519.
\newblock \href {http://arxiv.org/abs/1805.11519} {\path{arXiv:1805.11519}}.
\newline\urlprefix\url{http://arxiv.org/abs/1805.11519}

\bibitem{wang2016studying}
Z.~Wang, S.~Chang, Y.~Yang, D.~Liu, T.~S. Huang, Studying very low resolution
  recognition using deep networks, in: Proceedings of the IEEE Conference on
  Computer Vision and Pattern Recognition, 2016, pp. 4792--4800.

\bibitem{zou2011very}
W.~W. Zou, P.~C. Yuen, Very low resolution face recognition problem, IEEE
  Transactions on image processing 21~(1) (2011) 327--340.

\bibitem{Lu2018dcr}
Z.~{Lu}, X.~{Jiang}, A.~{Kot}, Deep coupled resnet for low-resolution face
  recognition, IEEE Signal Processing Letters 25~(4) (2018) 526--530.

\bibitem{Distillation_2018Low}
S.~Ge, S.~Zhao, C.~Li, J.~Li, Low-resolution face recognition in the wild via
  selective knowledge distillation, IEEE Transactions on Image Processing
  28~(4) (2018) 2051--2062.

\bibitem{lr_kd1}
S.~Zhao, X.~Gao, S.~Li, S.~Ge, Low-resolution face recognition in the wild with
  mixed-domain distillation, in: 2019 IEEE Fifth International Conference on
  Multimedia Big Data (BigMM), IEEE, 2019, pp. 148--154.

\bibitem{tip2020-ge2020efficient}
S.~Ge, S.~Zhao, C.~Li, Y.~Zhang, J.~Li, Efficient low-resolution face
  recognition via bridge distillation, TIP 29 (2020) 6898--6908.

\bibitem{ddl2020}
Y.~Huang, P.~Shen, Y.~Tai, S.~Li, X.~Liu, J.~Li, F.~Huang, R.~Ji, Improving
  face recognition from hard samples via distribution distillation loss, in:
  ECCV, Springer, 2020, pp. 138--154.

\bibitem{ge2020look}
S.~Ge, K.~Zhang, H.~Liu, Y.~Hua, S.~Zhao, X.~Jin, H.~Wen, Look one and more:
  Distilling hybrid order relational knowledge for cross-resolution image
  recognition, in: Proceedings of the AAAI Conference on Artificial
  Intelligence, Vol.~34, 2020, pp. 10845--10852.

\bibitem{cao2018vggface2}
Q.~Cao, L.~Shen, W.~Xie, O.~M. Parkhi, A.~Zisserman, Vggface2: A dataset for
  recognising faces across pose and age, in: 2018 13th IEEE international
  conference on automatic face \& gesture recognition (FG 2018), IEEE, 2018,
  pp. 67--74.

\bibitem{rpcl_xu1993rival}
L.~Xu, A.~Krzyzak, E.~Oja, Rival penalized competitive learning for clustering
  analysis, rbf net, and curve detection, IEEE Transactions on Neural networks
  4~(4) (1993) 636--649.

\bibitem{xu2007scholarRPCL}
L.~Xu, Rival penalized competitive learning, Scholarpedia 2~(8) (2007) 1810.

\bibitem{DeepID2+_sun2015deeply}
Y.~Sun, X.~Wang, X.~Tang, Deeply learned face representations are sparse,
  selective, and robust, in: Proceedings of the IEEE conference on computer
  vision and pattern recognition, 2015, pp. 2892--2900.

\bibitem{DeepID3_sun2015deepid3}
Y.~Sun, D.~Liang, X.~Wang, X.~Tang, Deepid3: Face recognition with very deep
  neural networks, arXiv preprint arXiv:1502.00873.

\bibitem{LFW_Gary2008}
G.~B. Huang, M.~Mattar, T.~Berg, E.~Learned-Miller, Labeled faces in the wild:
  A database forstudying face recognition in unconstrained environments, 2008.

\bibitem{lr_explore_aghdam2019exploring}
O.~A. Aghdam, B.~Bozorgtabar, H.~K. Ekenel, J.-P. Thiran, Exploring factors for
  improving low resolution face recognition, in: 2019 IEEE/CVF Conference on
  Computer Vision and Pattern Recognition Workshops (CVPRW), IEEE, 2019, pp.
  2363--2370.

\bibitem{liPei2019low}
P.~Li, L.~Prieto, D.~Mery, P.~J. Flynn, On low-resolution face recognition in
  the wild: Comparisons and new techniques, IEEE Transactions on Information
  Forensics and Security 14~(8) (2019) 2000--2012.

\bibitem{ren2012coupled}
C.-X. Ren, D.-Q. Dai, H.~Yan, Coupled kernel embedding for low-resolution face
  image recognition, IEEE Transactions on Image Processing 21~(8) (2012)
  3770--3783.

\bibitem{biswas2011multidimensional}
S.~Biswas, K.~W. Bowyer, P.~J. Flynn, Multidimensional scaling for matching
  low-resolution face images, IEEE transactions on pattern analysis and machine
  intelligence 34~(10) (2011) 2019--2030.

\bibitem{Fine-to-coarse_peng2016fine}
X.~Peng, J.~Hoffman, X.~Y. Stella, K.~Saenko, Fine-to-coarse knowledge transfer
  for low-res image classification, in: 2016 IEEE International Conference on
  Image Processing (ICIP), IEEE, 2016, pp. 3683--3687.

\bibitem{casia_yi2014learning}
D.~Yi, Z.~Lei, S.~Liao, S.~Z. Li, Learning face representation from scratch,
  arXiv preprint arXiv:1411.7923.

\bibitem{celeb_guo2016ms}
Y.~Guo, L.~Zhang, Y.~Hu, X.~He, J.~Gao, Ms-celeb-1m: A dataset and benchmark
  for large-scale face recognition, in: European conference on computer vision,
  Springer, 2016, pp. 87--102.

\bibitem{bulat2018learn}
A.~Bulat, J.~Yang, G.~Tzimiropoulos, To learn image super-resolution, use a gan
  to learn how to do image degradation first, in: Proceedings of the European
  conference on computer vision (ECCV), 2018, pp. 185--200.

\bibitem{scface_grgic2011scface}
M.~Grgic, K.~Delac, S.~Grgic, Scface--surveillance cameras face database,
  Multimedia tools and applications 51~(3) (2011) 863--879.

\bibitem{cheng2018tinyface}
Z.~Cheng, X.~Zhu, S.~Gong, Low-resolution face recognition, in: Asian
  Conference on Computer Vision, Springer, 2018, pp. 605--621.

\bibitem{YTF_Wolf2011Face}
L.~Wolf, T.~Hassner, I.~Maoz, Face recognition in unconstrained videos with
  matched background similarity, in: CVPR 2011, IEEE, 2011, pp. 529--534.

\bibitem{megaface_kemelmacher2016megaface}
I.~Kemelmacher-Shlizerman, S.~M. Seitz, D.~Miller, E.~Brossard, The megaface
  benchmark: 1 million faces for recognition at scale, in: Proceedings of the
  IEEE conference on computer vision and pattern recognition, 2016, pp.
  4873--4882.

\bibitem{MDS_mudunuri2015low}
S.~P. Mudunuri, S.~Biswas, Low resolution face recognition across variations in
  pose and illumination, IEEE transactions on pattern analysis and machine
  intelligence 38~(5) (2015) 1034--1040.

\bibitem{DMDS_LDMDS_yang2017}
F.~Yang, W.~Yang, R.~Gao, Q.~Liao, Discriminative multidimensional scaling for
  low-resolution face recognition, IEEE Signal Processing Letters 25~(3) (2017)
  388--392.

\bibitem{RICNN_zeng2016}
D.~Zeng, H.~Chen, Q.~Zhao, Towards resolution invariant face recognition in
  uncontrolled scenarios, in: 2016 International Conference on Biometrics
  (ICB), IEEE, 2016, pp. 1--8.

\bibitem{LightCNN_wu2018light}
X.~Wu, R.~He, Z.~Sun, T.~Tan, A light cnn for deep face representation with
  noisy labels, IEEE Transactions on Information Forensics and Security 13~(11)
  (2018) 2884--2896.

\bibitem{VGGFace_parkhi2015deep}
O.~M. Parkhi, A.~Vedaldi, A.~Zisserman, Deep face recognition.

\bibitem{tcn2019-zha2019tcn}
J.~Zha, H.~Chao, Tcn: Transferable coupled network for cross-resolution face
  recognition, in: ICASSP, IEEE, 2019, pp. 3302--3306.

\bibitem{discriminative_yang2017}
F.~Yang, W.~Yang, R.~Gao, Q.~Liao, Discriminative multidimensional scaling for
  low-resolution face recognition, IEEE Signal Processing Letters 25~(3) (2017)
  388--392.

\bibitem{mm2019-ge2019fewer}
S.~Ge, S.~Zhao, X.~Gao, J.~Li, Fewer-shots and lower-resolutions: Towards
  ultrafast face recognition in the wild, in: ACM MM, 2019, pp. 229--237.

\end{thebibliography}

\end{document}